# A Multivariate Discretization Method for Learning Bayesian Networks from Mixed Data

**Stefano Monti**[†]
[†]Intelligent Systems Program
University of Pittsburgh
901M CL, Pittsburgh, PA – 15260
smonti@isp.pitt.edu

**Gregory F. Cooper**[†‡]
[‡]Center for Biomedical Informatics
University of Pittsburgh
8084 Forbes Tower, Pittsburgh, PA – 15261
gfc@cbmi.upmc.edu

## Abstract

In this paper we address the problem of discretization in the context of learning Bayesian networks (BNs) from data containing both continuous and discrete variables. We describe a new technique for *multivariate* discretization, whereby each continuous variable is discretized while taking into account its interaction with the other variables. The technique is based on the use of a Bayesian scoring metric that scores the discretization policy for a continuous variable given a BN structure and the observed data. Since the metric is relative to the BN structure currently being evaluated, the discretization of a variable needs to be dynamically adjusted as the BN structure changes.

## 1 Introduction

In most approaches to learning Bayesian networks (BNs) from data, simplifying assumptions are made to circumvent practical problems in the implementation of the theory. One common assumption is that all variables are discrete [3, 7], or that all variables are continuous and normally distributed [6]. This is often too restrictive an assumption, since most real world domains are best described by a combination of continuous and discrete attributes (mixed data).

A possible solution around this limitation is to discretize the continuous variables, so as to be able to apply one of the well established techniques for learning BNs containing discrete variables only. The alternative is the direct modeling of the continuous data as such. However, when the direct modeling is based on a limited set of families of probability densities, it can fail to capture the nature of the interaction between the continuous variables. The adoption of a more general parametric representation is marred by a often prohibitive computational cost. In these cases, discretization can actually be preferable, and some empirical evidence supports this conclusion [4, 12]. One of the appeals of this method is that discretization can be interpreted as a form of non-parametric density estimation [14]. An additional advantage of this approach is its relative simplicity. However, discretization can in general generate spurious dependencies among variables, especially discretization strategies that deal with the single variables individually, without considering their interaction with other variables. We refer to this class of discretization strategies as *univariate* discretization, to distinguish it from *multivariate* discretization, whereby each variable is discretized taking into account its interaction with the other variables.

The majority of the discretization techniques currently available are devised with a classification task in mind. In current state-of-the-art discretization methods for classification, the search for the best discretization of a given feature variable is carried out by at most considering its interaction with the class variable of interest, while ignoring possible interactions with other feature variables [4]. For the reasons previously outlined, this approach does not seem appropriate for learning BNs, where modeling the interaction among *all* variables often is of primary importance, when for example the goal of the learning task is to gain insight into the causal or probabilistic dependencies that exist among the domain variables. Very few multivariate discretization strategies have been proposed in machine learning [2], and to our knowledge there is only one example of multivariate discretization strategy tailored to the needs of BN learning [5].

Similar to the approach proposed in [5], we describe a technique for multivariate discretization where each continuous variable is discretized by taking into consideration its interaction with the other variables. The technique is based on the use of a Bayesian scoring metric that scores the discretization policy of a continuous variable given the BN structure. In order to define the scoring metric according to the Bayesian paradigm, we make the discretization strategy part of the model generating the observed data. That is, we assume that there is an underlying discrete mechanism governing the behavior of the observed continuous variables. This framing of the discretization problem allows for the natural specification of a Bayesian scoring metric for discretization, defined to be the posterior probability of a discretization policy, given a BN structure and the data.

Since the proposed scoring metric for discretization is dependent on the BN structure currently searched, the discretization of a variable needs to be dynamically adjusted as the BN structure changes. The dependence of one variable's discretization on the other variables in the BN implies that the effects of the discretization will possibly propagate throughout the network.

The derived Bayesian scoring metric shares many of the properties of the scoring metric described in [5], which is based on the MDL principle. The adoption of the Bayesian paradigm, however, forces us to make explicit many modeling assumptions that could be left implicit in the MDL-based approach, and it provides for a possible semantics of the discretization process.

The remainder of this paper is organized as follows. In Section 2, we briefly introduce the Bayesian network formalism, and a particular method for learning BNs from data containing discrete variables only. In Section 3, we describe our approach to discretization. We first introduce it in the context of discretizing a single variable (univariate discretization). We then generalize it to the multivariate case. Finally, Section 4 lists a set of issues for future research.

## 2  Notation and background

In this section, we very briefly introduce the Bayesian network formalism. This is by no means a comprehensive introduction to the topic, and its main purpose is to establish some common notation, and to frame the research issue which is the focus of this paper.

In general, we denote random variables with upper case letters, such as $X$, $Y$, and their instantiation or realization with the corresponding lower case letters, $x$, $y$, or $x^{(l)}$, $y^{(l)}$, where we use the latter notation when we need to distinguish between different instantiations. Similarly, we denote random vectors with bold upper case letters, such as $\boldsymbol{V}, \boldsymbol{W}$, and their instantiation or realization with the corresponding bold lower case letters, $\boldsymbol{v}, \boldsymbol{w}$. Given a domain of interest, we denote with $\mathcal{X} = \{X_1, \ldots, X_n\}$ the complete set of variables in that domain, and with $\boldsymbol{x}$ or $\boldsymbol{x}^{(l)}$ the full instantiations of the variables in $\mathcal{X}$.

Marginal and conditional probabilities over arbitrary subsets $\boldsymbol{Y}$ and $\boldsymbol{Z}$ of $\mathcal{X}$ will be denoted with $\rho(\boldsymbol{Y})$ and $\rho(\boldsymbol{Y} \mid \boldsymbol{Z})$ respectively, and they denote either a probability density or a probability mass, depending on whether the variables involved are continuous or discrete.

### 2.1  Bayesian networks

A Bayesian network $B$ is defined by a pair $(S, \Theta_S)$, where $S = (\mathcal{X}, E)$ is a directed acyclic graph (DAG) with set of nodes $\mathcal{X}$, and with a set of arcs $E = \{(X_i, X_j) \mid X_i, X_j \in \mathcal{X}, X_i \neq X_j\}$ representing probabilistic dependencies among domain variables[1]. $\Theta_s$ represents the parameterization of a probability measure $\rho$ defined over the space of possible instantiations of $\mathcal{X}$. Given a node $X_i \in \mathcal{X}$, we use $\mathbf{Pa}_i$ to denote the set of parents of $X_i$ in $S$.

The essential property of BNs is summarized by the *Markov property*, which asserts that each variable is independent of its non-descendants given its parents [13]. This property allows for the representation of the multivariate joint probability distribution over $\mathcal{X}$ in terms of the univariate conditional distributions $\rho(X_i \mid \mathbf{Pa}_i, \Theta_S)$ of each variable $X_i$ given its parents $\mathbf{Pa}_i$. Application of the chain rule, together with the Markov property, yields the following factorization of the joint probability of any particular instantiation $\boldsymbol{x}$ of all $n$ variables:

$$\rho(\boldsymbol{x}) = \rho(x_1, \ldots, x_n) = \prod_{i=1}^{n} \rho(x_i \mid \mathbf{pa}_i, \Theta_S) \cdot \quad (1)$$

The complete set of conditional independence assertions implied by a network structure can be determined by means of the concept of *d-separation*, a graphical characterization introduced by Pearl [13]. Using the concept of d-separation, it is also possible to show that a BN variable, conditioned on the set of nodes containing its parents, its children, and its children's parents, is independent of all the other variables in the network. This set of nodes is called the *Markov blanket* of the variable.

---

[1]In this paper, we make no distinction between the network nodes and the variables they represent.

## 2.2 Learning Bayesian networks: A Bayesian approach

The task of learning BNs involves learning the network structure and learning the parameters of the conditional probability distributions. A well established set of learning methods is based on the definition of a scoring metric measuring the fitness of a network structure to the data, and on the search for high-scoring network structures based on the defined scoring metric [3, 7, 9].

In a Bayesian framework, ideally classification and prediction would be performed by taking a weighted average over the inferences of every possible BN containing the domain variables. Since this method is in general computationally infeasible, often an attempt has been made to use a high scoring BN for classification and prediction. The latter approach is also preferable when the goal of the learning task is to gain insight into the causal or statistical dependencies that may exist among the domain variables. We will assume the use of this approach in the remainder of this paper.

The basic idea of the Bayesian approach is to maximize the probability $\rho(S \mid \mathcal{D}) = \rho(S, \mathcal{D})/\rho(D)$ of a network structure $S$ given a database of cases $\mathcal{D}$, and a set of assumptions and priors that we leave implicit. Because for all network structures the term $\rho(\mathcal{D})$ is the same, for the purpose of model selection it suffices to calculate $\rho(S, \mathcal{D}) = \rho(\mathcal{D} \mid S)\rho(S)$.

The term $\rho(S)$ is the prior probability of the structure S, and needs to be given as input. The term $\rho(\mathcal{D} \mid S)$ is the *marginal likelihood*, also called the *evidence*, and it measures how well the given structures fits the data.

The computation of the marginal likelihood involves the evaluation of a high-dimensional integral. The analytical evaluation of this integral has been developed for the cases when all variables in $\mathcal{X}$ are discrete, [3, 7] or all the variables are continuous and normally distributed [6]. In the general case however, and in particular when dealing with data containing both continuous and discrete variables, it is unlikely that analytic evaluation of the term $\rho(\mathcal{D} \mid S)$ is feasible, and approximations would need to be used [1].[2]

A relatively simple alternative is to discretize the continuous data, so as to be able to apply one of the scoring metrics available for discrete domains. This is the approach investigated in this paper, and the remainder of this section is devoted to the description of the Bayesian scoring metric for discrete domains upon which our discretization strategy is built. To simplify exposition, in the remainder of this paper we will assume a uniform prior $\rho(S)$ over the space of BN structures.

### 2.2.1 A Bayesian scoring metric for discrete domains

For each variable $X_i \in \mathcal{X}$, let $r_i$ indicate the number of values $X_i$ can take, and let $q_i$ indicate the number of distinct values its parent set $\mathbf{Pa}_i$ can take. With these notational conventions, let the set of parameters $\Theta_S$ be decomposable as:

$$\begin{aligned}
\Theta_S &= \{\boldsymbol{\theta}_1, \ldots, \boldsymbol{\theta}_n\} \\
\boldsymbol{\theta}_i &= \{\boldsymbol{\theta}_{i1}, \ldots, \boldsymbol{\theta}_{iq_i}\}, \quad i = 1, \ldots, n \\
\boldsymbol{\theta}_{ij} &= \{\theta_{ij1}, \ldots, \theta_{ijr_i}\}, \quad j = 1, \ldots, q_i.
\end{aligned} \quad (2)$$

As it should be clear from the definition of a BN, for each node $X_i$, the vector $\boldsymbol{\theta}_i$ is the set of parameters necessary to fully characterize the conditional probability distribution $\rho(X_i \mid \mathbf{Pa}_i)$. Accordingly, the parameter set $\boldsymbol{\theta}_{ij}$ specifies the distribution of $X_i$ conditioned on the $j$-th instantiation of the parent set $\mathbf{Pa}_i$. Finally, the parameter $\theta_{ijk} = \rho(X_i = k \mid \mathbf{Pa}_i = j)$ specifies the probability of observing the $k$-th value of $X_i$ conditioned on the observation of the $j$-th value of $\mathbf{Pa}_i$.

Given the above decomposition of the parameter vector $\Theta_S$, if we further assume Dirichlet priors over $\Theta_S$, of the form $\boldsymbol{\theta}_{ij} \sim \text{Dirichlet}(\alpha_{ij1}, \ldots, \alpha_{ijr_i})$, for all $X_i$'s and for all instantiations of $\mathbf{Pa}_i$, then we can factorize the prior probability of the parameter set $\Theta_S$ as follows:

$$\rho(\Theta_S \mid S) \propto \prod_{i=1}^{n} \prod_{j=1}^{q_i} \prod_{k=1}^{r_i} \theta_{ijk}^{\alpha_{ijk}-1}. \quad (3)$$

The assumption of Dirichlet priors is a strong but very convenient assumption, since it implies strong assumptions of parameter independence within the set $\Theta_S$, as is evident from the form of Equation (3).[3] Under the assumptions just described, and provided $\mathcal{D}$ has no missing data, the likelihood $\rho(\mathcal{D} \mid S)$ can be evaluated analytically, and, expressed in log terms, it has the following form:

$$\begin{aligned}\log \rho(\mathcal{D} \mid S) &= \sum_{i=1}^{n} \left[ \sum_{j=1}^{q_i} \log \frac{\Gamma(\alpha_{ij})}{\Gamma(\alpha_{ij} + N_{ij})} \right. \\ &\quad + \left. \sum_{k=1}^{r_i} \log \frac{\Gamma(\alpha_{ijk} + N_{ijk})}{\Gamma(\alpha_{ijk})} \right] \\ &\equiv \sum_{i=1}^{n} \mathcal{S}_d(X_i, \mathbf{Pa}_i; \mathcal{D}), \qquad (4)\end{aligned}$$

where $\Gamma(\cdot)$ is the Gamma function[4], and $\alpha_{ij} = \sum_k \alpha_{ijk}$, with the $\alpha_{ijk}$ as part of the Dirichlet prior specification. Also, $N_{ijk}$ is the number of cases in $\mathcal{D}$ where the variable $X_i = k$, and the parent set $\mathbf{Pa}_i = j$, and $N_{ij}$ is the number of cases in $\mathcal{D}$ where $X_i$'s parent set $\mathbf{Pa}_i$ takes its $j$-th value, irrespective of the value of $X_i$ [3, 7].

---
[2] A notable exception is the *Conditional Gaussian* model introduced by Lauritzen in [11], whereby a mixture of continuous and discrete variables is modeled by assuming that each continuous variable has no discrete descendants.

[3] In [7] it is shown that there is a close relationship between these assumptions and a Dirichlet prior for $\Theta_S$.

In Equation (4), we use the term $\mathcal{S}_d(X_i, \mathbf{Pa}_i; \mathcal{D})$ to denote the group of terms between square brackets. This notation clearly shows the *decomposability* of the likelihood term, in that the overall score is given by a sum of terms $\mathcal{S}_d(X_i, \mathbf{Pa}_i; \mathcal{D})$, each measuring the contribution of a node and its parents. The subscript $d$ in $\mathcal{S}_d$ is to emphasize that the score is defined over discrete variables only. This qualification will become relevant in the next section, where the score defined here will be one of the building blocks for our discretization strategy.

## 3 Bayesian discretization

The discretization of a continuous variable can be interpreted as the selection of the number of values the discretized variable should take, as well as of the thresholds in the continuous range of the continuous variable that delimit the intervals to be mapped into the values of the discretized variable. Given a dataset $\mathcal{D}$, the number of values the discretized variable can take is upper-bounded by the cardinality $N$ of $\mathcal{D}$. Furthermore, as candidate thresholds for the partition of the continuous range of a continuous variable, we only consider the (at most) $N - 1$ mid-points between contiguous data points in $\mathcal{D}$.

Let $\Lambda_{X_i} = \{r_i, \Delta_{X_i}\}$ be a *discretization policy* for the continuous variable $X_i$, where $r_i \geq 2$ is the number of categories used in the discretization of $X_i$, and $\Delta_{X_i} = \{\delta_{i1}, \ldots, \delta_{ir_i-1}\}$ is the set of thresholds in the value range of $X_i$ delimiting these categories.[5] We denote with $Y_i$ the discretized variable corresponding to $X_i$, taking values $\{y_i^1, \ldots, y_i^{r_i}\}$. Variable $Y_i$'s relationship to $X_i$ is fully specified by the mapping $Y_i = y(X_i)_\Lambda$ (the subscript $\Lambda$ in $y(x_i)_\Lambda$ will often be dropped as the discretization policy used will be clear from the context), defined as:

$$y(X_i)_\Lambda = \begin{cases} y_i^1 & \text{if } X_i \leq \delta_1, \\ y_i^{r_i} & \text{if } X_i > \delta_{r_i-1}, \\ y_i^k & \text{if } \delta_{k-1} < X_i \leq \delta_k, \\ & \quad k = 2, \ldots, r_i - 1. \end{cases} \qquad (5)$$

We denote the interval $(\delta_{ik-1}, \delta_{ik}]$ with $[y_i^k]$, with the boundary cases to be interpreted as $[y_i^1] \equiv (-\infty, \delta_{i1}]$ and $[y_i^{r_i}] \equiv (\delta_{ir_i-1}, +\infty)$. Finally, we denote with $\Lambda = \{\Lambda_1, \ldots, \Lambda_n\}$ the discretization policy for the whole set of continuous variable in $\mathcal{X}$, and with $y(\mathcal{X})_\Lambda = \{y(X_1)_{\Lambda_1}, \ldots, y(X_n)_{\Lambda_n}\}$ the discretization of all the variables in $\mathcal{X}$ according to the discretization policy $\Lambda$. To simplify exposition, we consider a discretization policy over the whole set $\mathcal{X}$, and we assume the trivial discretization $y(X_i) = X_i$ for each discrete variable $X_i$ in X.

The specification of a discretization strategy involves the definition of a scoring metric $\mathcal{S}(\Lambda; \mathcal{D}, S)$, which scores a discretization policy with respect to a database $\mathcal{D}$ and a BN structure $S$, and the definition of a search algorithm to search for the discretization policy that maximizes the given scoring metric.[6]

Our approach to the specification of a multivariate discretization strategy is based on a problem transformation, whereby we suppose that each of the continuous variables in the domain of interest is a noisy observation of an underlying discrete variable. The problem of finding a "good" discretization is thus translated into the problem of finding a set of discrete variables, and their (probabilistic) mapping into the corresponding continuous variables, that best account for the observed interactions between the continuous variables.

In the remainder of this section, we show that this framing of the discretization problem allows for the natural specification of a Bayesian scoring metric for discretization policies. We first illustrate our approach when applied to a single variable. We then generalize it to the multivariate case.

---

[4]For an integer value $n > 0$, the Gamma function corresponds to the factorial function offset by one, i.e., $\Gamma(n+1) = n!$.

[5]These thresholds are such that $\delta_{i1} < \ldots < \delta_{ir_i-1}$.

[6]In keeping with the usual notation for score functions, we denote a generic scoring metric with $\mathcal{S}(\alpha; \beta)$ where the left parameter(s) $\alpha$ is the parameter with respect to which we want to maximize the score, while the right parameter(s) $\beta$ is kept constant.

### 3.1 Univariate Bayesian discretization

In the univariate case, we have $\mathcal{X} = \{X\}$, with $X$ a continuous variable, and the dataset $\mathcal{D} = \boldsymbol{x} = \{x^{(1)}, \ldots, x^{(N)}\}$. As previously pointed out, the purpose of the discretization is the computation of the marginal likelihood $\rho(\boldsymbol{x} \mid S)$. Conditioning on $S$ is in this case superfluous, since we are dealing with a one-variable BN, therefore $\rho(\boldsymbol{x} \mid S) = \rho(\boldsymbol{x})$.

As informally outlined in the previous section, the problem of scoring a given discretization policy can be formalized by assuming that the mechanism by which the data $\boldsymbol{x}$ was generated by the environment involves a discrete variable. More specifically, we assume that the density of the observed continuous variable $X$ is governed by an underlying discrete variable $Y$, whose relationship to $X$ is fully specified by the (unknown) discretization policy $\Lambda$, and the (unknown) $\Lambda$-*induced* conditional density function $\rho(X \mid Y, \Lambda, \boldsymbol{\theta}_\Lambda)$. We assume that $Y$ follows a multinomial distribution with a Dirichlet prior of the form described in Section 2.2.1. Furthermore, although the parameters $\boldsymbol{\theta}_\Lambda$ of the conditional density function $\rho(X \mid Y, \Lambda, \boldsymbol{\theta}_\Lambda)$ will be in general unknown, we restrict its general form to be as shown in Table 1, where $1_{\{\cdot\}}$ is the indicator function (i.e., $1_{\{cond\}} = 1$, if *cond* holds, 0 otherwise), and it is used to bound the support of the density to be within the appropriate interval.

Table 1: Conditional probability density

| $Y$ | $\rho(X \mid Y, \Lambda, \boldsymbol{\theta}_\Lambda)$ |
|---|---|
| $y^1$ | $\rho(X \mid y^1, \Lambda, \boldsymbol{\theta}_{\Lambda 1}) 1_{\{X \in [y^1]\}}$ |
| ... | ... |
| $y^r$ | $\rho(X \mid y^r, \Lambda, \boldsymbol{\theta}_{\Lambda r}) 1_{\{X \in [y^r]\}}$ |

We further assume that all the probability components are of the same parametric form and that the corresponding parameters are independent, that is, that the prior $\rho(\boldsymbol{\theta}_\Lambda)$ can be factorized as $\rho(\boldsymbol{\theta}_\Lambda) = \prod_{k=1}^{r} \rho(\boldsymbol{\theta}_{\Lambda k})$.

To fully specify a probabilistic model, we also need to give a probabilistic meaning to a discretization policy $\Lambda$. To this end, and to simplify notation, we use $\Lambda$ to denote both the discretization policy and the corresponding *hypothesis* that $\Lambda$ is the discretization policy according to which the data $\boldsymbol{x}$ were generated. Furthermore, we assume that all the possible discretization policies of the form $\Lambda$ represent a mu-tually exclusive and exhaustive set of hypotheses.[7] This allows for the specification of a prior probability $\rho(\Lambda)$ over the space of possible discretization policies $\{\Lambda\}$.

Based on this conceptualization of the discretization problem, we now show that an appropriate score $\mathcal{S}(\Lambda; \mathcal{D})$ for a discretization policy $\Lambda$ with respect to the data $\mathcal{D}$ is given by the (logarithm of) the posterior probability of $\Lambda$ given the data $\mathcal{D}$. In fact, we can rewrite the marginal likelihood $\rho(\boldsymbol{x})$ by making explicit its dependency on $Y$ and $\Lambda$,

$$\begin{aligned}
\rho(\boldsymbol{x}) &= \sum_\Lambda \sum_{\boldsymbol{y}} \rho(\boldsymbol{x} \mid \boldsymbol{y}, \Lambda) \rho(\boldsymbol{y} \mid \Lambda) \rho(\Lambda) \\
&= \sum_\Lambda \rho(\boldsymbol{x} \mid \boldsymbol{y}(\boldsymbol{x}), \Lambda) \rho(\boldsymbol{y}(\boldsymbol{x}) \mid \Lambda) \rho(\Lambda) \\
&\equiv \sum_\Lambda \rho(\boldsymbol{x} \mid \Lambda) \rho(\Lambda) ,
\end{aligned} \quad (6)$$

where the elimination of the summation over $\boldsymbol{y}$ is possible because of the particular nature of the probability function $\rho(X \mid Y, \Lambda)$, which assigns probability zero whenever $\boldsymbol{y} \neq \boldsymbol{y}(\boldsymbol{x})$.

From Equation (6), we see that a possible solution to its computation is to sample from the $\Lambda$ space a given number of data points, and use model averaging to approximate the likelihood. Alternatively, application of the *maximum a posteriori* (MAP), also known as *plug-in*, approximation to solve Equation (6) yields

$$\rho(\boldsymbol{x}) \simeq \rho(\boldsymbol{x} \mid \tilde{\Lambda}) = \rho(\boldsymbol{x} \mid \boldsymbol{y}(\boldsymbol{x}), \tilde{\Lambda}) \rho(\boldsymbol{y}(\boldsymbol{x}) \mid \tilde{\Lambda}) , \quad (7)$$

where $\tilde{\Lambda}$ is the posterior mode of $\Lambda$, i.e., $\tilde{\Lambda} = \operatorname{argmax}_\Lambda [\rho(\Lambda \mid \boldsymbol{x})]$. The selection of a discretization policy is thus reduced to the problem of finding the discretization policy that maximizes the posterior probability $\rho(\Lambda \mid \boldsymbol{x})$ or, given the proportionality $\rho(\Lambda \mid \boldsymbol{x}) \propto \rho(\boldsymbol{x}, \Lambda)$, the discretization policy that maximizes the joint probability $\rho(\boldsymbol{x}, \Lambda) = \rho(\boldsymbol{x} \mid \Lambda) \rho(\Lambda)$. Furthermore, if we assume a uniform prior probability $\rho(\Lambda)$, the problem reduces to maximum likelihood estimation, i.e., to the search for the discretization policy that maximizes the likelihood $\rho(\boldsymbol{x} \mid \Lambda)$ or, equivalently, the log-likelihood $\log \rho(\boldsymbol{x} \mid \Lambda)$.

The computation of $\rho(\boldsymbol{y}(\boldsymbol{x}) \mid \tilde{\Lambda})$ in Equation (7) is straightforward. Based on the previously made assumption that $\boldsymbol{y}(\boldsymbol{x})$ is a multinomial sample with Dirichlet prior, we notice that $\log \rho(\boldsymbol{y}(\boldsymbol{x}) \mid \Lambda)$ is a special case of the log-likelihood of Equation (4). In this case, we have a one-variable BN, given by the variable $Y$ with an empty parent set. We can thus apply the result established in Equation (4) to the computation of $\log \rho(y(x) \mid \Lambda)$ to obtain

$$\log \rho(y(x) \mid \Lambda) = \mathcal{S}_d(Y, \emptyset, \Lambda; \mathcal{D}) =$$
$$\log \frac{\Gamma(\alpha)}{\Gamma(\alpha + N)} + \sum_{k=1}^{r} \log \frac{\Gamma(\alpha_k + N_k)}{\Gamma(\alpha_k)}, \quad (8)$$

where $N_k$ denotes the number of cases with $y(X) = y^k$, the $\alpha_k$'s are part of the prior specification of $\theta_Y$, and $\alpha = \sum_k \alpha_k$. Notice that in a slight change of notation from Section 2.2.1, we have here made explicit the dependency on $\Lambda$ in $\mathcal{S}_d(Y, \emptyset, \Lambda; \mathcal{D})$. The dependency of the above expression on the discretization policy $\Lambda$ is through the sufficient statistics $N_k$, whose value depends on how the continuous range of $X$ is partitioned by $\Lambda$.

---

[7]Provided we consider as discretization thresholds only the mid-points between the data points in $\boldsymbol{x}$, this set is finite, and it has cardinality $\sum_{l=1}^{N-1} \binom{N-1}{l} = 2^{(N-1)}$.

For the computation of the term $\rho(x \mid y(x), \Lambda)$ of Equation (7), we need to specify the parametric form of the conditional density $\rho(X \mid Y, \Lambda)$. A convenient choice is to assume that within each interval $[y^k]$, the values of $X$ are distributed uniformly, i.e.,

$$\rho(x \mid y^k, \Lambda) = \frac{dx}{\delta_k - \delta_{k-1}} 1_{\{x \in [y^k]\}} \equiv \rho_k, \quad (9)$$

where $\rho_k$ is a constant for any given $\Lambda$. Notice that the above distribution is not normalizable for the boundary cases corresponding to the intervals $[y^1]$ and $[y^r]$. We can easily solve this problem by using the smallest value of $X$ in the dataset as a lower-bound for the interval $[y^1]$, and the largest value of $X$ in the dataset as an upper-bound for the interval $[y^r]$. Alternatively, for those variables whose domain is known to be bound above and below (a situation frequently occurring, e.g., with variables measuring medical findings), we can use these domain-specific bounds to delimit the two intervals $[y^1]$ and $[y^r]$.

The use of the uniform distribution considerably simplifies the computation of $\rho(x \mid y(x), \Lambda)$., since the probability of each case $x^{(l)}$ is only dependent on $y(x^{(l)})$. In fact, if we denote with $x_l = \{x^{(1)}, \ldots, x^{(l-1)}\}$ the set of the first $l - 1$ cases in $x$, we can rewrite $\rho(x \mid y(x), \Lambda)$ as follows:

$$\rho(x \mid y(x), \Lambda) = \prod_{l=1}^{N} \rho(x^{(l)} \mid y(x^{(l)}), x_l, y(x_l), \Lambda)$$
$$= \prod_{l=1}^{N} \rho(x^{(l)} \mid y(x^{(l)}), \Lambda) = \prod_{k=1}^{r} \rho_k^{N_k}, \quad (10)$$

which, in log terms, yields $\log \rho(x \mid y(x), \Lambda) = \sum_{k=1}^{r} N_k \log \rho_k \equiv \mathcal{S}_c(X, Y, \Lambda; \mathcal{D})$.

To summarize, under the assumption of a uniform prior $\rho(\Lambda)$, we define the scoring metric for a univariate discretization policy $\Lambda$, given the data $\mathcal{D} = x$ as

$$\mathcal{S}(\Lambda; \mathcal{D}) = \mathcal{S}_d(Y, \emptyset, \Lambda; \mathcal{D}) + \mathcal{S}_c(X, Y, \Lambda; \mathcal{D}), \quad (11)$$

where we refer to the term $\mathcal{S}_d(Y, \emptyset, \Lambda; \mathcal{D})$ of Equation (8), as the *discrete component* of the score, and to the term $\mathcal{S}_c(X, Y, \Lambda; \mathcal{D})$ of Equation (10), as the *continuous component* of the score.

The continuous and the discrete components in Equation (11) play complementary roles in the determination of the score of a given discretization policy. On the one hand, the continuous component rewards prediction accuracy with respect to the continuous variable (i.e., it measures how well the discretized data predict the original non-discretized data). It also rewards model complexity. This should be clear by looking at Equation (10), where we expressed the continuous component of the score as a function of the $\rho_k$'s, the parameters specifying the uniform density. As the width of an interval $[y^k]$ becomes smaller, the corresponding $\rho_k$ increases, thus increasing the probability of any given point $x$ in that interval. As a consequence, everything else being equal, increasing the number $r$ of partition of a variable will in general increase the continuous component of the corresponding score, which could lead to overfitting the data. On the other hand, the discrete component penalizes model complexity. In fact, this component measures the likelihood of the discretized data, and this increases as we decrease the number of values of the discretized variable. In the limit case of a one-value variable, the discrete component will be highest, since that one value will always have probability 1 (with reference to Equation (8), for $r = 1$ the discrete component of the score takes the maximum attainable value of 0, since $N_k = N$, and the two terms of the sum cancel out). As a result, the scoring metric of Equation (11) tries to establish a trade-off between model accuracy and model complexity.

### 3.1.1 Choice of the conditional density

In the previous section, we used the uniform distribution for the conditional density $\rho(X \mid Y, \Lambda)$ of a continuous variable $X$ given its discretized counterpart $Y$. Choices other than the uniform distribution are possible. However, the resulting conditional likelihood $\rho(x \mid y(x), \Lambda)$ would not have as simple a form (because of the indicator function in Table 1, the given distribution would be truncated). Parametric distributions belonging to the exponential family seem to be the best choice in light of the fact that they allow for the analytical computation of the relevant statistics.

The discretization metric described above can be applied to the case where we want to "discretize" an already discrete variable, by grouping together subsets of its values, a transformation more appropriately characterized as *abstraction*. In this case, an appropriate choice would be to model the conditional $\rho(X \mid Y, \Lambda)$ as a multinomial distribution with a Dirichlet prior, which would result in a continuous component of the score having the same form as the discrete component of the score.

We can use the multinomial distribution as a proxy for the conditional density also in those cases when $X$ is continuous. Provided we choose a uniform prior, the resulting density would "default" to a uniform distribution in those cases where a given interval contains datapoints that are all distinct. However, in those intervals with repetitions, the conditional distribution would give higher probability to the repeated values. This is very similar to the conditional distribution implied by the MDL-metric described in [5]. In fact, in a MLD framework, each continuous value within a given interval is assigned a Huffman code whose length is inversely proportional to the frequency of occurrence of that value within the interval. This code is approximately equal to $-\log f_x$, where $f_x$ is the frequency of occurrence of the value $x$. When there are no repetitions within an interval, each value is assigned an equal length code, which corresponds to assuming a uniform distribution within that interval.

### 3.1.2 Choice of the prior probability

In this section, we briefly explore the use of informative priors for $\rho(\Lambda)$, and how these could be specified. As previously illustrated, the discretization of a single variable $X$ is defined as $\Lambda = \{r, \Delta_X\}$, with $r$ the number of categories of $Y = y(X)$, and $\Delta_X$ the set of cutpoints in the continuous domain of $X$. A natural way of factorizing the prior $\rho(\Lambda_X)$ is thus in terms of its components, that is, $\rho(\Lambda_X) = \rho(\Delta_X \mid r)\rho(r)$.

An informative prior over $r$ is relatively simple to specify. For example, we might model $\rho(r)$ as a truncated Poisson distribution (truncated above at $N-1$, where $N$ is the number of cases in the database), with mean $\lambda$, $2 \leq \lambda \leq N - 1$. The Poisson distribution is a sensible choice in light of the fact that it penalizes large values of $r$, which would tend to be favored if we based our selection on the likelihood term $\rho(x \mid \Lambda)$ only.

Given $r$, a uniform prior probability $\rho(\Delta_X \mid r)$ could be assumed. Alternatively an informative prior could be defined based on the definition of a "baseline" discretization, and of a "distance" metric measuring the difference of a given discretization from this baseline. The baseline could be based on simple discretization method (e.g., equal bin discretization), or it could be provided by a domain expert.

### 3.2 Multivariate Bayesian discretization

We can generalize the approach illustrated in the previous section to the multivariate case, where $\mathcal{X} = \{X_1, \ldots, X_n\}$, with $n > 1$. As in the univariate case, we assume the existence of an underlying discrete mechanism that governs the behavior of the observed continuous variables in $\mathcal{X}$. That is, we assume the existence of a set of discrete variables $\mathcal{Y} = \{Y_1, \ldots, Y_n\}$, where each variable $Y_i$ governs the behavior of the corresponding continuous variable $X_i$, and that the relationship between each $X_i$ and the corresponding $Y_i$ is fully specified by the discretization policy $\Lambda = \{\Lambda_1, \ldots, \Lambda_n\}$ and the set of $\Lambda$-induced conditional densities $\{\rho(X_i \mid Y_i, \Lambda_i, S)\}_{i=1,\ldots,n}$. Notice that in this case the conditioning on $S$ cannot be ignored. In fact, to complete the specification of the discrete mechanism, we further assume that the probabilistic dependencies over $\mathcal{X}$ implied by the structure $S$ are in fact probabilistic dependencies over the discrete variables $\mathcal{Y}$ which are manifested over $\mathcal{X}$ when the variables in $\mathcal{Y}$ are marginalized out. This assumption corresponds to transforming the BN structure $S$ into a structure $S'$ augmented with the variables in $\mathcal{Y}$, such that each continuous variable $X_i$ in $S'$ has $Y_i = y(X_i)$ as its unique parent, and the parent set of $Y_i$ in $S'$ contains $Y_j = y(X_j)$ if and only if $X_j$ is a parent of $X_i$ in $S$. In graphical terms, the introduction of the discrete variables $\mathcal{Y}$ corresponds to the transformation shown in Figure 1. Notice that an immediate consequence of this transformation is that a continuous variable $X_i$ is independent of its parents $\mathbf{Pa}_i$ given its corresponding discretized variable $Y_i$, i.e., $\forall X_i \in \mathcal{X}$,

$$\rho(X_i \mid y(X_i), \mathbf{Pa}_i, \Lambda) = \rho(X_i \mid y(X_i), \Lambda) . \quad (12)$$

Equation (12) simply asserts that the discretized value of a variable $X_i$ is all we need to know to compute the probability of its non-discretized value. An important consequence of this result is that the conditional probability of a complete case $x$ given its discretization $y(x)$ is decomposable as follows:

$$\rho(x \mid y(x), \Lambda) = \prod_{X_i \in \mathcal{X}} \rho(x_i \mid y(x_i), \Lambda) . \quad (13)$$

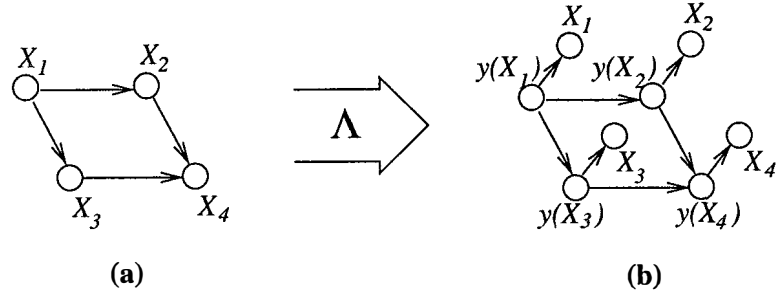

Figure 1: Graphical interpretation of the introduction of the discrete mechanism: a) the graph modeling the probabilistic dependencies between the continuous variables $X_1$ to $X_4$; and b) the graph with discrete variables $y(X_1)$ to $y(X_4)$ assumed to model the observed interactions between variables $X_1$ to $X_4$.

Based on Equation (13), on the application of the Markov property (as defined in Equation (1), Section 2.1), and on the assumption that the prior $\rho(\Lambda \mid S)$ is decomposable, i.e., $\rho(\Lambda \mid S) = \prod_i \rho(\Lambda_i \mid S)$, we can factorize the joint probability of a case $x$ as follows:

$$\rho(x \mid \Lambda) = \left[ \prod_{X_i \in \mathcal{X}} \rho(x_i \mid y(x_i), \Lambda_i) \right] \times \left[ \prod_{X_i \in \mathcal{X}} \rho(y(x_i) \mid y(\mathbf{pa}_i), \Lambda_i) \right]. \quad (14)$$

From Equation (14), we can see that the joint probability of a database case $x$ containing both continuous and discrete variables can be factorized as a product of terms containing discrete variables only (i.e., the product over the terms $\rho(y(x_i) \mid y(\mathbf{pa}_i), \Lambda_i)$) times a product of terms containing continuous variables only (i.e., the product over the terms $\rho(x_i \mid y(x_i), \Lambda_i)$).

We thus have all the ingredients for the definition of the scoring metric of a multivariate discretization policy $\Lambda$, given a BN structure $S$ and database $\mathcal{D}$. Analogously to the univariate case, under the assumption of a uninformative prior, this metric is taken to be the log-likelihood $\log \rho(\mathcal{D} \mid \Lambda, S)$ of the data given the discretization policy $\Lambda$, and the BN structure $S$ (if we use an informative prior, we just need to add the extra term $\log \rho(\Lambda)$).

The metric can be computed as follows. Let $\Lambda_i = \{\Lambda_i\} \cup \{\Lambda_j : X_j \in \mathbf{Pa}_i\}$ be the set of discretization policies for $X_i$ and its parent set $\mathbf{Pa}_i$. Also, let $y(\mathbf{Pa}_i)$ denote the set of discrete variables associated with the continuous variables in $\mathbf{Pa}_i$. Finally, let $\mathcal{D}_l = \{x^{(1)}, \ldots, x^{(l-1)}\}$ be the set of first $l-1$ cases in the database $\mathcal{D}$. Then, the scoring metric for multivariate discretization $\mathcal{S}(\Lambda; \mathcal{D}, S)$ can be defined as

$$\mathcal{S}(\Lambda; \mathcal{D}, S) = \sum_{i=1}^{n} \mathcal{S}_c(X_i, y(X_i), \Lambda_i; \mathcal{D}) + \sum_{i=1}^{n} \mathcal{S}_d(y(X_i), y(\mathbf{Pa}_i), \Lambda_i; \mathcal{D}). \quad (15)$$

The terms $\mathcal{S}_c(X_i, y(X_i), \Lambda_i; \mathcal{D})$ in Equation (15) are the continuous components of the score, and under the assumption of a uniform distribution for the conditional densities $\rho(X_i \mid y(X_i), \Lambda_i)$, they are defined as shown in Equation (10).

The terms $\mathcal{S}_d(y(X_i), y(\mathbf{Pa}_i), \Lambda_i; \mathcal{D})$ in Equation (15) are the discrete components of the score, and they are defined as the analogous terms $\mathcal{S}_d(X_i, \mathbf{Pa}_i; \mathcal{D})$ of Equation (4), which were based on the variables $X_i$ and $\mathbf{Pa}_i$ being discrete.

With regard to the discrete components, it is important to emphasize the dependency on $\Lambda_i$ of each discrete component $\mathcal{S}_d(y(X_i), y(\mathbf{Pa}_i), \Lambda_i; \mathcal{D})$. This dependency is manifested through the sufficient statistics $N_{ij}$ and $N_{ijk}$, which are now based on discretized variables. As such, they are dependent on the discretization policy. More importantly, the sufficient statistics for a given $\mathcal{S}_d(y(X_i), y(\mathbf{Pa}_i), \Lambda_i; \mathcal{D})$ term depend on the discretization policy of $X_i$ as well as on the discretization policy of the variables in $\mathbf{Pa}_i$. The immediate consequence of this dependency is that when selecting the discretization policy $\Lambda_i$ for a given variable $X_i$ with respect to the BN structure S, we should aim to maximize the following *local score*:

$$\mathcal{S}(\Lambda_i; \mathcal{D}, S) = \mathcal{S}_c(X_i, y(X_i), \Lambda_i; \mathcal{D}) + \mathcal{S}_d(y(X_i), y(\mathbf{Pa}_i), \Lambda_i; \mathcal{D}) \quad (16)$$
$$+ \sum_{X_j \in \mathbf{Ch}_i} \mathcal{S}_d(y(X_j), y(\mathbf{Pa}_j), \Lambda_j; \mathcal{D}),$$

where $\mathbf{Ch}_i$ is the set of children of $X_i$. In fact, each of the $\Lambda_j$ contains the discretization policy $\Lambda_i$. It follows that if we change $\Lambda_i$ (i.e, if we change the discretization of $X_i$), besides recomputing $\mathcal{S}_d(y(X_i), y(\mathbf{Pa}_i), \Lambda_i; \mathcal{D})$, we also need to recompute the term $\mathcal{S}_d(y(X_j), y(\mathbf{Pa}_j), \Lambda_j; \mathcal{D})$ for each of the $X_j$ (if any) having $X_i$ among its parents, i.e., for each of the $X_j$ in $\mathbf{Ch}_i$. As previously indicated, not surprisingly this result agrees with the results derived based on the MDL principle described in [5].

A consequence of Equation (16) is that changing the discretization of a variable $X_i$ will possibly affect the discretization of all the other variables in the network which are not d-separated from $X_i$ either by the empty set, or by some d-separator containing discrete variables only (e.g., if the BN contains continuous variables only, the discretization of any variable will possibly affect the discretization of all the other variables in the network).

### 3.3 Searching the space of discretization policies

Putting aside search strategies that search directly over the space of joint discretizations of all dependent variables at once, the simplest search strategy is to start from some initial discretization of all variables. We then search for the maximum score discretization for each single variable while keeping the discretization of the other variables fixed. We repeat this process until the overall score does not improve of a given amount (this is the search strategy adopted in [5]).

Here, we want to focus on a search strategy which is suggested by the similarity between the set of independency properties satisfied by the discretization score, and the probabilistic independencies implied by a BN structure. As outlined in the previous section, the discretization of a variable $X_i$ is independent of the discretization of a variable $X_j$, if the two variables are d-separated by the empty set or by a set containing discrete variables only.[8] It also follows that if we hold *fixed* the discretization of a variable d-separating $X_i$ and $X_j$, the discretization of these two variables can be carried out independently of each other's.

Notice that these are the same independence assertions used in the definition of efficient exact algorithms for BN inference. Based on these similarities, we can draw a natural correspondence between the problem of finding the most probable discretization of all the continuous variables in the BN and the problem of finding the most probable instantiation of all the variables of a BN. For this purpose, we need to replace the act of holding fixed the *instantiation* of a variable in the BN inference problem, with the act of holding fixed the *discretization* of a variable in the multivariate discretization problem. We can thus modify any of the available exact or approximate algorithms for BN inference that can return the most probable instantiation of the BN variables (e.g., [8, 10]), so as to have it return the most probable discretization instead.

---

[8] In this discussion, we refer to independence assertions implied by the original structure $S$, not the augmented structure $S'$.

While the method outlined above allows for a more efficient search over the space of multivariate discretizations, exact computation (i.e., selection of the optimal discretization) often remains infeasible. In fact, since even in the univariate case searching for the optimal discretization has a complexity that is exponential in the number of data points, it is clear that the multivariate search outlined above still has to rely on some form of heuristic search when selecting the thresholds to be included in the discretization of a given variable.

Another solution worth investigating is the application of Monte Carlo techniques applied to Equation (6).

## 4 Conclusions and future work

We have described a new approach to multivariate discretization that is specifically tailored to the task of learning Bayesian networks from data containing both continuous and discrete variables. The method is based on the specification of a Bayesian scoring metric for discretization policies. In order to derive the scoring metric, we have relied on a conceptualization of the discretization problem in terms of the existence of an underlying discrete mechanism governing the behavior of the observed continuous variables. We want to emphasize here that, although there are real domains for which such an interpretation could be entirely plausible,[9] we are not suggesting this model has causal plausibility in general. Rather, we view this conceptualization as the basis for an initial Bayesian analysis which is subject to modification based on future research. There are several issues that need to be further investigated. Extensive experimental evaluation need to be performed, in order to assess the merits of the proposed method. We are currently working on the implementation of a search algorithm for the selection of a high-scoring discretization policy, based on the adaptation of the clique-tree propagation algorithm for BN inference [8].

---

[9] As an example, think of a radio dial, which is defined on a continuous range, but that works only for a finite number of frequencies. As the dial is moved from the "right" location, the reception degrades, until a latent thresholds is crossed, when the reception shift to another frequency.

Another important issue that needs to be addressed regards the specification of informative priors over the space of discretization policies $\{\Lambda\}$. We pointed out that the specification of such a prior could be useful to avoid attaining too finely grained a discretization, which would tend to be preferred over "coarser" discretizations if the selection was based solely on the data.

Another issue we consider worth exploring is the use of density functions other than the uniform for the specification of the conditional density $\rho(X|Y,\Lambda)$. We believe a more informative density could yield more accurate discretizations.

## Acknowledgments

We thank Roger Day and the anonymous reviewers for their useful comments on a preliminary version of this manuscript. The research presented here was supported in part by grants BES-9315428 and IRI-9509792 from the National Science Foundation and by grant LM05291-02 from the National Library of Medicine.